\documentclass[10pt,twocolumn,letterpaper]{article}

\usepackage[pagenumbers]{cvpr} 
\usepackage{mmstyle}

\newcommand{\tablestyle}[2]{\setlength{\tabcolsep}{#1}\renewcommand{\arraystretch}{#2}\centering\footnotesize}

\usepackage[dvipsnames]{xcolor}

\newcommand\blfootnote[1]{%
  \begingroup
  \renewcommand\thefootnote{}\footnote{#1}%
  \addtocounter{footnote}{-1}%
  \endgroup
}
\usepackage{subcaption}

\newcommand{\first}{\includegraphics[width=10px]{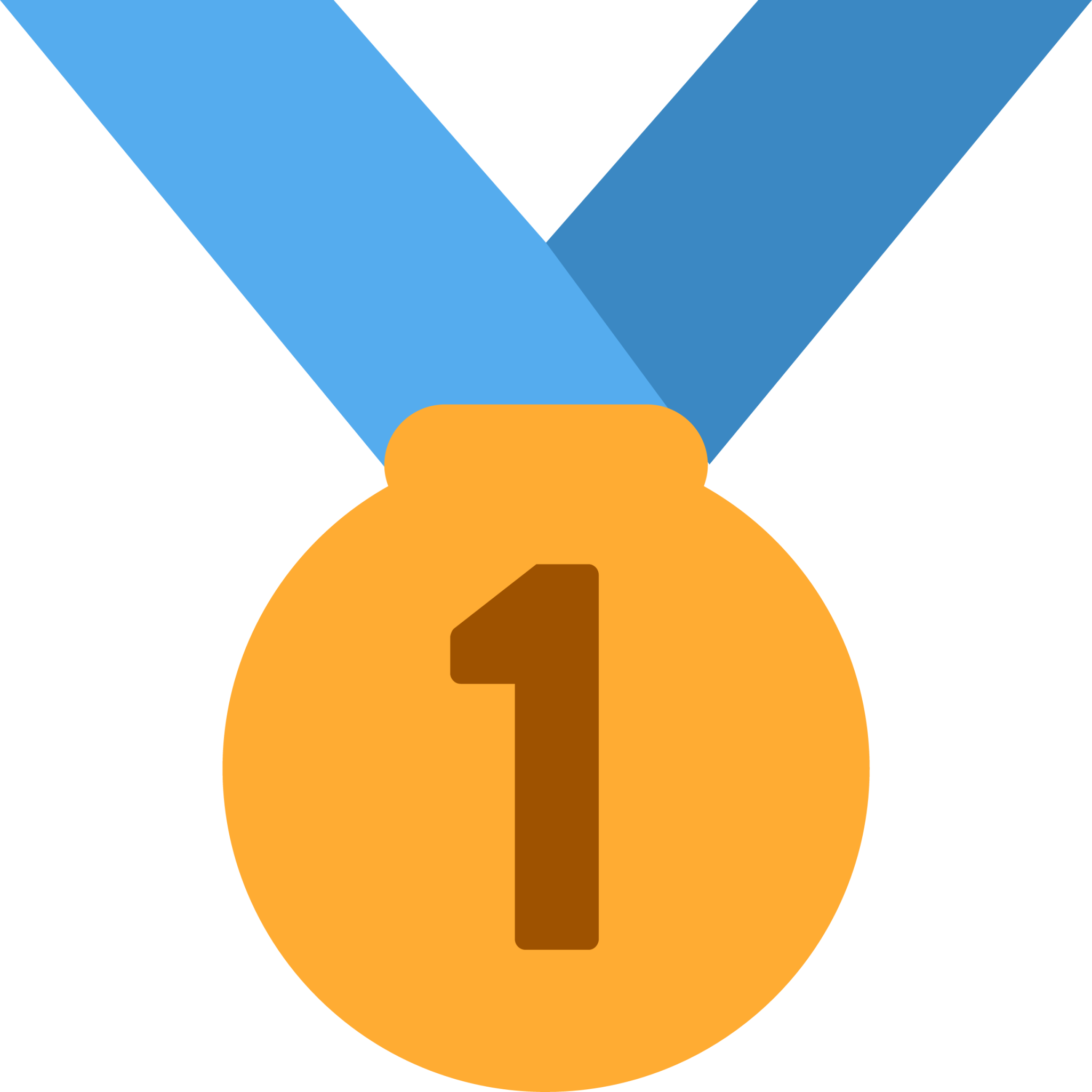}}
\newcommand{\second}{\includegraphics[width=10px]{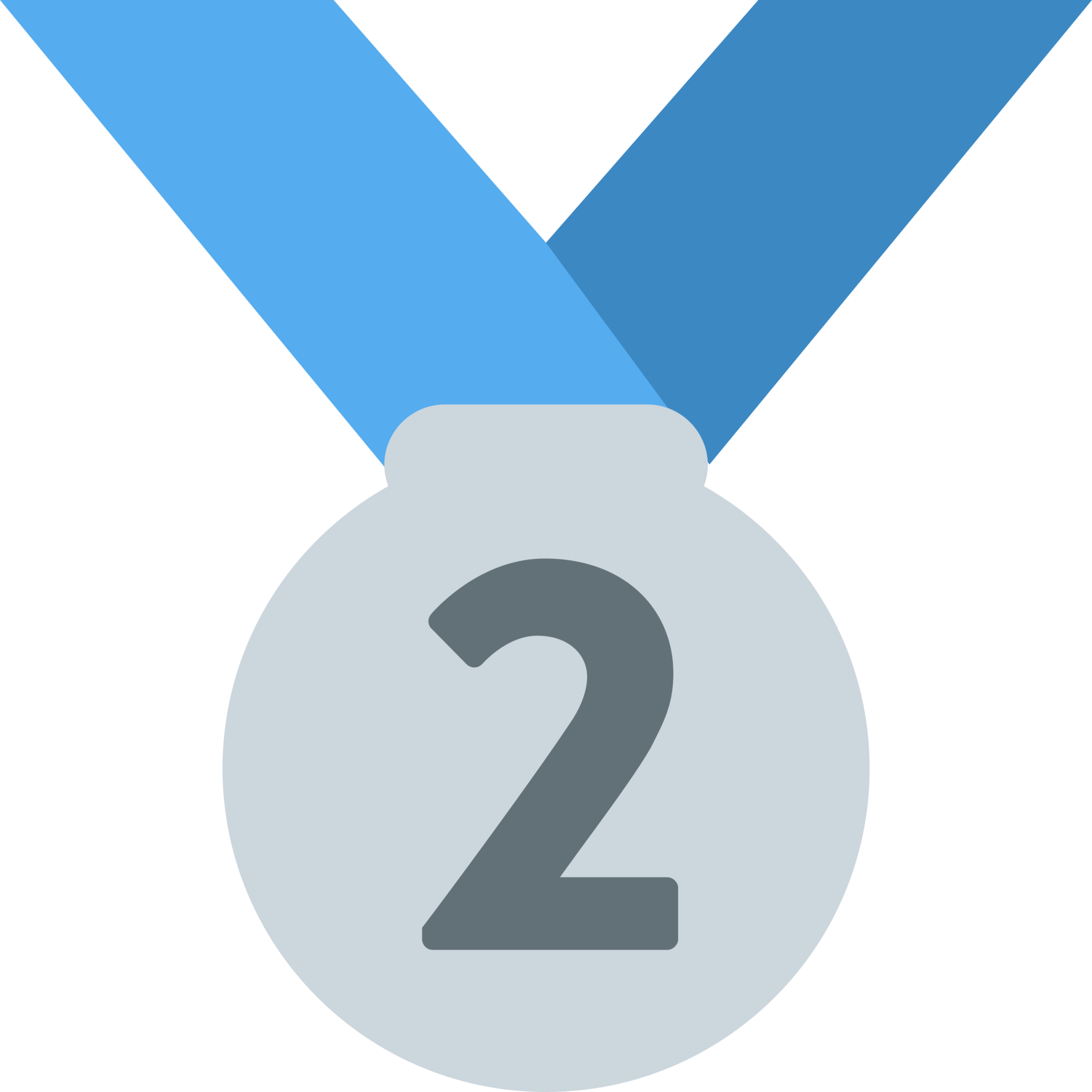}}
\newcommand{\third}{\includegraphics[width=10px]{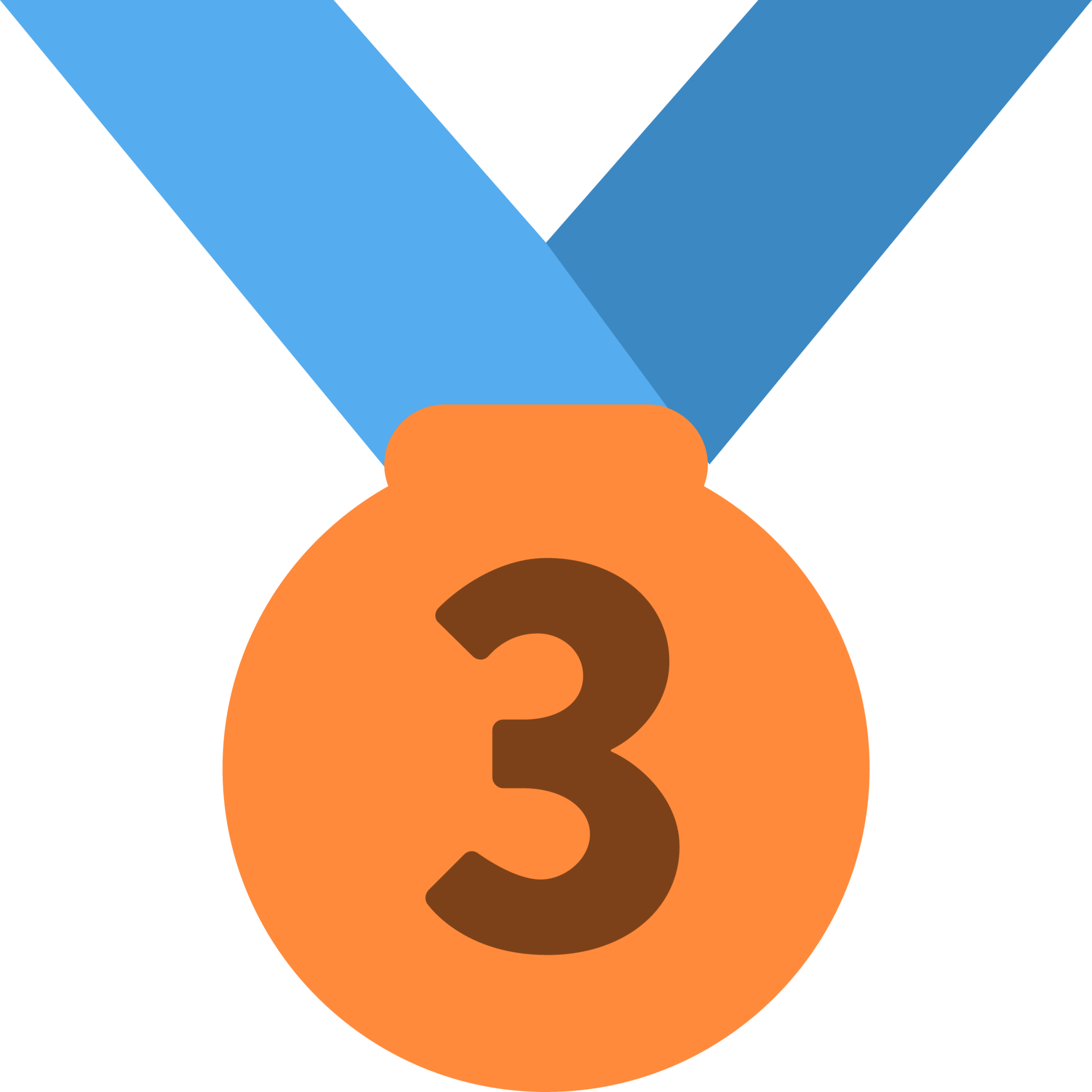}}

\definecolor{cvprblue}{rgb}{0.21,0.49,0.74}
\usepackage[pagebackref,breaklinks,colorlinks,citecolor=cvprblue]{hyperref}

\def\paperID{*****} 
\def\confName{CVPR}
\def\confYear{2024}

\title{V3Det Challenge 2024 on Vast Vocabulary and Open Vocabulary Object Detection: Methods and Results}

\author{
Jiaqi Wang, Yuhang Zang, Pan Zhang, Tao Chu, Yuhang Cao, Zeyi Sun, Ziyu Liu, Xiaoyi Dong,  \\
Tong Wu, Dahua Lin, Zeming Chen, Zhi Wang, Lingchen Meng, Wenhao Yao, Jianwei Yang, \\
Sihong Wu, Zhineng Chen, Zuxuan Wu, Yu-Gang Jiang, Peixi Wu, Bosong Chai, Xuan Nie, \\
Longquan Yan, Zeyu Wang, Qifan Zhou, Boning Wang, Jiaqi Huang, Zunnan Xu, Xiu Li \\
Kehong Yuan, Yanyan Zu, Jiayao Ha, Qiong Gao, Licheng Jiao
}

\begin{document}
\maketitle
\blfootnote{
Jiaqi Wang, Pan Zhang, Tao Chu, Yuhang Cao, Zeyi Sun, Ziyu Liu, Xiaoyi Dong, Yuhang Zang, Tong Wu, Dahua Lin are the organizers of the V3Det challenge, and other authors participated in the challenge.
The Appendix lists the authors’ teams and affiliations.}
\begin{abstract}
Detecting objects in real-world scenes is a complex task due to various challenges, including the vast range of object categories, and potential encounters with previously unknown or unseen objects.
The challenges necessitate the development of public benchmarks and challenges to advance the field of object detection.
Inspired by the success of previous COCO~\cite{lin2014microsoft} and LVIS~\cite{gupta2019lvis} Challenges, we organize the \textbf{V3Det Challenge 2024} in conjunction with the 4th Open World Vision Workshop: Visual Perception via Learning in an Open World (VPLOW) at CVPR 2024, Seattle, US. This challenge aims to push the boundaries of object detection research and encourage innovation in this field.
The V3Det Challenge 2024 consists of two tracks: 1) \textbf{Vast Vocabulary Object Detection}: This track focuses on detecting objects from a large set of $13204$ categories, testing the detection algorithm's ability to recognize and locate diverse objects. 2) \textbf{Open Vocabulary Object Detection}: This track goes a step further, requiring algorithms to detect objects from an open set of categories, including unknown objects.
In the following sections, we will provide a comprehensive summary and analysis of the solutions submitted by participants. By analyzing the methods and solutions presented, we aim to inspire future research directions in vast vocabulary and open-vocabulary object detection, driving progress in this field. Challenge homepage: \href{https://v3det.openxlab.org.cn/challenge}{https://v3det.openxlab.org.cn/challenge}.
\end{abstract}
\section{Introduction}
Object detection has witnessed tremendous advancements in recent years~\cite{girshick2014rich,ren2015faster,lin2017focal,lin2017feature,he2017mask,chen2019hybrid,chen2019mmdetection,tian2019fcos,wang2020seesaw,carion2020end}, with public benchmarks and challenges~\cite{everingham2010pascal,lin2014microsoft,gupta2019lvis,kuznetsova2020open} serving as a crucial catalyst for innovation.
However, the challenges in real-world object detection remain exists, such as dealing with vast vocabulary (\ie, large numbers of object classes) and open-vocabulary scenarios (\ie, detecting objects from unseen or unknown classes).
To contribute to the advancement of object detection methods to handle vast vocabulary and open-vocabulary scenarios, the V3Det Challenge 2024 has been established.
The V3Det challenge aims to foster innovation in robust and general object detection algorithms.

This challenge is based on the Vast Vocabulary Visual Detection (V3Det) dataset~\cite{wang2023v3det}, which not only encompasses objects from 13,204 categories, ten times the size of existing large vocabulary object detection datasets but also emphasizes the hierarchical and interrelated nature of categories, providing an ideal testbed for research in extensive and open vocabulary object detection.
The rich annotations of V3Det, meticulously provided by human experts, ensure high precision and in-depth interpretation of the data.

The V3Det Challenge 2024 features two tracks for participants.
The first track is vast vocabulary object detection with $13204$ categories, focusing on detecting objects with a large vocabulary of categories. The second track is open vocabulary object detection, which is split into base and novel classes, and only allows annotations of base classes for training.
The evaluation server is hosted on the Eval AI platform~\footnote{\href{https://eval.ai/web/challenges/challenge-page/2250/overview}{https://eval.ai/web/challenges/challenge-page/2250/overview}}. 
In total, 20 teams made valid submissions to the challenge tracks.
The methods of top teams on vast vocabulary and open vocabulary are presented in Sec.~\ref{sec:vast} and Sec.~\ref{sec:ovd}, respectively.
\begin{table*}[t!]
    \begin{minipage}{.49\linewidth}
    \centering
    \caption{\small{Results of top teams on the \textbf{vast vocabulary} object detection track.}}
    \vspace{-6pt}
    \label{tab:vast-main}
    \tablestyle{8pt}{1.2}
    \begin{tabular}{l|c|c|c|c|c|c}
    Team  & AP & AP$_{50}$ & AP$_{75}$ & AP$_{s}$ & AP$_{m}$ & AP$_{l}$ \\
    \hline
    \first czm369  & \textbf{54.4} & \textbf{60.7} & \textbf{56.7} & 20.7 & 33.5 & \textbf{60.8} \\
    \second lcmeng  & 53.1 & 59.0 & 55.5 & 23.7 & \textbf{35.9} & 58.9 \\
    \third TCSVT   & 50.6 & 56.4 & 52.8 & \textbf{24.3} & 34.6 & 55.8 \\
    JYYY    & 43.8 & 50.7 & 46.4 & 13.8 & 23.9 & 49.5
    \end{tabular}
    \end{minipage}
    \hspace{+1mm}
    \begin{minipage}{.49\linewidth}
    \centering
    \caption{\small{Results of top teams on the \textbf{open vocabulary} object detection track.}}
    \vspace{-6pt}
    \label{tab:ovd-main}
    \tablestyle{8pt}{1.2}
    \begin{tabular}{l|c|c|c}
    Team  & $\rm{AP^{final}}$ & $\rm{AP^{novel}}$ & $\rm{AP^{base}}$ \\
    \hline
    \first lcmeng          & \textbf{22.9} & \textbf{15.6} & 44.8 \\
    \second TCSVT          & 20.3          & 10.3          & \textbf{50.4} \\
    \third Innovision      & 14.2          & 4.5           & 43.4 \\
    Host            & 10.9          & 5.5           & 27.0 
    \end{tabular}
    \end{minipage}
\end{table*}

\section{Vast Vocabulary Object Detection Track}\label{sec:vast}
The vast vocabulary object detection track evaluates models of supervised learning for object detection on all 13204 classes of the V3Det dataset. Public academic datasets such as object detection, image classification datasets (except for the Bamboo~\cite{zhang2022bamboo}), and image-text caption datasets are all permitted to be used in this track.

\paragraph{Evaluation Metrics} The results are evaluated on the test set of V3Det with $29863$ images. The mAP metric is adopted following the COCO dataset~\cite{lin2014microsoft}. A maximum of 300 boxes per image are allowed in the evaluation.

\paragraph{Results} Table~\ref{tab:vast-main} showcases the results of selected top teams on the vast vocabulary track. We will describe their detailed technical solution in the following sections.

\subsection{Solution of First Place (czm369)}
\noindent \textbf{MixPL v2: Expanding Detection Categories via Semi-Supervised Learning}

\paragraph{Motivation} In Table~\ref{tab:first_vast}, they provide an in-depth analysis of the importance of using unlabeled data with the number of categories increasing. Fine-tuning the EVA model on the COCO and LVIS datasets achieves significant detection performance improvements. However, when finetuning on the V3Det dataset, the gains are more modest. This indicates that fine-tuning base models struggle to effectively map pre-trained visual representations to the nuanced semantics of significantly expanded detection categories. In contrast, semi-supervised learning effectively bridges the gap between high-granularity category semantics and pre-trained visual representations through the use of pseudo-labels. This enables the motivation of combining self-training with vast vocabulary object detection.

\paragraph{Pipeline} The winning team designs a semi-supervised pipeline MixPL v2 utilizing MixPL~\cite{chen2023mixed} that uses the V3Det as the labeled dataset and the Objects365~\cite{shao2019objects365} as the unlabeled dataset.
Based on the Mean Teacher~\cite{tarvainen2017mean}, they design the Mixed Pseudo Label pipeline for semi-supervised object detection.
They apply weak and strong augmentations for unlabeled images and use the teacher model to predict pseudo-labels on weakly augmented images.
The teacher model is the accumulated Exponential Moving Average (EMA) of the student model.
They select the Mixup~\cite{zhang2017mixup} and Mosaic~\cite{bochkovskiy2020yolov4} as augmentation techniques.

\paragraph{Implementation Details} They apply the designed semi-supervised learning algorithms on the Co-DETR~\cite{zong2023detrs} model and Swin-L~\cite{liu2021swin} backbone. They adopt additional techniques such as Gradient Accumulation for loss computation to reduce the negative impact of small batch size.

\begin{table*}[t!]
    \begin{minipage}{.5\linewidth}
    \centering
    \caption{\footnotesize{Ablation studies of the 1st-place solution in the \textit{vast vocabulary} track: the performance discrepancy between fine-tuning of base models and semi-supervised learning for category expansion.}}
    \vspace{-6pt}
    \label{tab:first_vast}
    \scalebox{.99}{
    \tablestyle{2pt}{1.}
    \begin{tabular}{l|l|l|l|l|l}
    Method  & Dataset & Classes & AP & AP$_{50}$ & AP$_{75}$ \\
    \hline
    EVA~\cite{fang2023eva} & COCO & 80 & 64.2 & 81.9 & 70.6 \\
    EVA~\cite{fang2023eva} & LVIS & 1203 & 62.2 & 76.2 & 65.4 \\
    EVA~\cite{fang2023eva} & V3Det & 13204 & 49.4 & 54.8 & 51.4 \\
    MixPL~\cite{chen2023mixed} & V3Det & 13204 & 54.5 & 60.6 & 56.7 \\ 
    \end{tabular}}
    \vspace{-6pt}
    \end{minipage}
    \hspace{+2mm}
    \begin{minipage}{.45\linewidth}
    \centering
    \caption{\footnotesize{Main results of the 2nd-place solution in the \textit{vast vocabulary} track.}}
    \vspace{-6pt}
    \label{tab:second_vast}
    \tablestyle{4pt}{1.}
    \begin{tabular}{l|c|c|c}
    Team  & AP & AP$_{50}$ & AP$_{75}$ \\
    \hline
    V3Det-DeformDETR-SwinB~\cite{wang2023v3det}  & 43.6 & - & - \\
    V3Det-DINO-SwinB~\cite{wang2023v3det}        & 43.1 & - & - \\
    V3Det-CascadeRCNN-EVA-Huge~\cite{wang2023v3det} & 50.7 & - & - \\
    \hline
    RichSem-DINO-Focal-Huge & 53.0 & 58.6 & 55.3 \\
    +TTA & 53.1 & 59.0 & 55.5 \\
    \end{tabular}
    \vspace{-6pt}
    \end{minipage}
\end{table*}

\begin{table*}[t!]
    \begin{minipage}{.3\linewidth}
    \centering
    \caption{\footnotesize{Ablation studies of the 3rd-place solution in the \textit{vast vocabulary} track.}}
    \vspace{-6pt}
    \label{tab:third_vast}
    \scalebox{.99}{
    \tablestyle{4pt}{1.}
    \begin{tabular}{l|c|c}
    Method & AP & AR \\
    \hline
    Baseline & 43.4 & 64.3 \\
    +PA-FPN~\cite{liu2018path} & 42.2 & 62.6 \\
    +DIoU~\cite{zheng2020distance} & 44.7 & 69.3 \\
    +GFL~\cite{li2021generalized} & 43.7 & 68.3 \\
    \end{tabular}}
    \end{minipage}
    \hspace{+1mm}
    \begin{minipage}{.3\linewidth}
    \centering
    \caption{\footnotesize{Ablation studies of the 4th-place solution in the \textit{vast vocabulary} track.}}
    \vspace{-6pt}
    \label{tab:fourth_vast}
    \scalebox{.99}{
    \tablestyle{4pt}{1.}
    \begin{tabular}{l|c}
    Method & AP \\
    \hline
    DINO~\cite{zhang2022dino} & 42.5  \\
    YOLOv8~\cite{varghese2024yolov8} & 21.0 \\
    DETR~\cite{carion2020end} & 40.0 \\
    Cascade R-CNN~\cite{cai2018cascade} & 44.0 \\
    \end{tabular}}
    \end{minipage}
    \hspace{+1mm}
    \begin{minipage}{.33\linewidth}
    \centering
    \caption{\footnotesize{Main results of the 1st-place solution in the \textit{open vocabulary} track.}}
    \vspace{-6pt}
    \label{tab:first_ovd}
    \scalebox{.99}{
    \tablestyle{2pt}{1.}
    \begin{tabular}{l|c|c|c}
    Team  & $\rm{AP}$ & $\rm{AP^{n}}$ & $\rm{AP^{b}}$ \\
    \hline
    CenterNet2-RN50-OVD~\cite{zhou2022detecting} & 9.7 & 3.3 & 28.6 \\
    Detic-RN50-ImageNet~\cite{zhou2022detecting} & 11.5 & 5.8 & 28.6  \\
    RichSem-DINO-Focal-Huge-TTA & 22.9 & 15.6 & 44.8 \\
    \end{tabular}}
    \end{minipage}
\end{table*}

\subsection{Solution of Second Place (lcmeng)}

\noindent \textbf{RichSem-DINO-FocalNet for V3Det Challenge 2024}

\paragraph{Pipeline} The second-place team, lcmeng, utilizes the RichSem-DINO~\cite{meng2024learning} detection framework, incorporating the FocalNet-Huge~\cite{yang2022focal} as its baseline. 
For better initialization, the model is firstly pre-trained on Objects365 dataset~\cite{shao2019objects365}, and then fine-tuned on the V3Det training set.

\paragraph{Implementation Details} The system is based on a two-stage Deformable DETR~\cite{zhu2020deformable} structure that uses multi-scale deformable attention for better high-resolution features. Concretely, the author uses 5 scale features and 900 object queries.
For further improvement, they use Test Time Augmentation (TTA) with two scale inputs and class-aware NMS (threshold 0.7) to filter redundant predictions.
Table~\ref{tab:second_vast} presents the main results of the 2nd place solution, which surpasses the previous SoTA by more than 2 AP under the supervised setting with the help pf TTA.

\subsection{Solution of Third Place (TCSVT)}

\noindent \textbf{Enhanced Object Detection: A Study on Vast Vocabulary Object Detection Track for V3Det Challenge 2024}

\paragraph{Pipeline} The third-place team, TCSVT~\cite{wu2024enhanced}, utilizes the Cascade R-CNN~\cite{cai2018cascade} detection framework, incorporating the Swin-B~\cite{liu2021swin} as its backbone. To better capture semantic information, they integrated a bottom-up structure inspired by PA-Net~\cite{liu2018path} into the Cascade R-CNN, enhancing shallow feature transmission and utilization.

\paragraph{Implementation Details} For better initialization, the Swin-B backbone is pretrained on the ImageNet-22K~\cite{russakovsky2014imagenet} dataset. To improve the training dataset's size and quality, the author applied data augmentations like flipping, jittering, and scaling. They also introduced the DIoU Loss~\cite{zheng2020distance} to address coordinate interrelationships and improve regression accuracy and convergence speed. Additionally, the Generalized Focal Loss (GFL)~\cite{li2021generalized} was applied to the Region Proposal Network to balance positive and negative sample proportions.

Table~\ref{tab:third_vast} shows the ablations results on the V3Det \textit{validation} set. They try different modules and use the DIoU loss and GFL loss in the final solution.

\subsection{Solution of Fourth Place (JYYY)}

\noindent \textbf{Team JYYY: V3Det Challenge 2024 - Vast Vocabulary Visual Detection}

The fourth-place team uses the ensemble solution of four detectors: DINO~\cite{zhang2022dino}, YOLO V8~\cite{varghese2024yolov8}, DETR~\cite{carion2020end}, and Cascade R-CNN~\cite{cai2018cascade}. 
Table~\ref{tab:fourth_vast} presents the ablation studies of different models, where Cascade R-CNN performs best among all the models.
They also apply post-processing techniques such as Test-Time Augmentation~\cite{shanmugam2021better} and Weighted Boxes Fusion~\cite{solovyev2021weighted}.

\subsection{Discussion}
Most participants in the vast vocabulary object detection track modify existing detectors with additional techniques like semi-supervised learning and transfer learning.
They use a strong detector pre-trained on massive data and fine-tune it on the V3Det dataset. However, this approach overlooks the domain gap between the pre-trained and target data. Participants also neglect the hierarchical relationships between categories. We believe there is still room for improvement.
\section{Open Vocabulary Object Detection}\label{sec:ovd}
This track splits all V3Det classes into the base (6709) and novel (6495) classes. The object detectors are expected to accurately detect objects of both base classes, with complete annotations, and novel classes, with only information including class names, class descriptions, and object-centric exemplar images for each class during inference.

\paragraph{Data}
For base classes, complete annotations are given. For novel classes, only class information is given, including class names, and class descriptions. Any other public datasets, if they don't include bounding box annotations for novel classes, such as other detection, image classification (except the Bamboo~\cite{zhang2022bamboo}), and text-image datasets are all permitted.

\paragraph{Evaluation Metrics}
The results will be evaluated on all images (29863) in the test set of V3Det for all 13204 categories. The results will be evaluated at most 300 boxes per image. The mAP metric is adopted as the metric following the COCO dataset. The performance of base and novel classes will be summarized as $\rm{AP^{base}}$ and $\rm{AP^{novel}}$. The final results for challenge ranking will be \begin{equation}
    \rm{AP^{final}} = \frac{\rm{AP^{base}} + 3 \cdot \rm{AP^{novel}}}{4}.
\end{equation}

\subsection{Results of the Open Vocabulary Track}
Table~\ref{tab:ovd-main} presents the results of the top teams. Below, we provide a brief overview of each team's framework. For team lcmeng and TCSVT, we only highlight the specialized design for open-vocabulary detection. Please refer to Sec.~\ref{sec:vast} for their detailed framework.

\subsection{Solution of the First Place (lcmeng)}

\noindent \textbf{RichSem-DINO-FocalNet for V3Det Challenge 2024}


To enable open-vocabulary detection, the author transformed the conventional closed-set classifier into alignment with prototype embeddings of both text and image samples. Specifically, they employed the Long-CLIP~\cite{zhang2024long} text encoder to extract text embeddings from GPT-4V~\cite{achiam2023gpt} category descriptions and used the CLIP-ViT-Large~\cite{radford2021learning} vision encoder to extract image embeddings from example images, averaging these embeddings for each class. 
To increase the number of example images per category, they implemented two strategies: cropping sub-images based on annotation bounding boxes in the training set, and using images from overlapping classes between V3Det and ImageNet-22K~\cite{russakovsky2014imagenet} as supplements.

Table~\ref{tab:first_ovd} represents the main results in the OVD track. The first-place solution achieves more
than 10 overall AP improvements compared to the baseline model even without introducing additional classification datasets into training.

\subsection{Solution of the Second Place (TCSVT)}
\label{sec:tcsvt-ovd}

\noindent \textbf{Enhanced Object Detection: A Study on Vast Vocabulary Object Detection Track for V3Det Challenge 2024}

To enable open-vocabulary detection, the author replaces the conventional closed-set classifier with the CLIP classifier used in Detic~\cite{zhou2022detecting}. Specifically, the CLIP~\cite{radford2021learning} text encoder extracts text embeddings from V3Det category descriptions, replacing the classifier's weights with these embeddings. The model is first trained on the V3Det base class training set, then directly infers results on the test set using the text embeddings of both base and novel classes.

\subsection{Solution of the Third Place (Innovision)}

\noindent \textbf{3rd Place Solution for OVD Track in CVPR 2024 VPLOW workshop: Visual Perception via Learning in an Open World}

The third-place team, Innovision, utilized the CenterNet2~\cite{zhou2021probabilistic} detection framework, incorporating the Swin Transformer~\cite{liu2021swin} as its backbone. CenterNet2 detects objects by identifying their centers and regressing to box parameters. Innovision enhanced this framework to operate at multiple scales using a Feature Pyramid Network (FPN)~\cite{lin2017feature}, generating feature maps with strides from 8 to 128 (P3-P7). Both classification and regression branches were applied at all FPN levels to produce detection heatmaps and bounding box regression maps.

For open-vocabulary detection, the author employs a similar strategy with team TCSVT, please refer the Sec.~\ref{sec:tcsvt-ovd} for details.

\subsection{Discussion}
In the open-vocabulary object detection track, participants use prior knowledge from Vision-Language Models such as CLIP~\cite{radford2021learning} and Long CLIP~\cite{zhang2024long} to recognize novel objects.
Despite this progress, the best AP in novel classes is only $15.6$ and there remains room for improvement.
Specifically, we encourage future research to explore combining Large Vision Language Models (LVLMs) to enhance recognition of novel objects~\cite{liu2024rar}.
\section{Conclusion}
This challenge aims to provide valuable insights into designing object detectors for vast and open vocabulary settings in real-world scenarios. We are pleased to see participants submit promising solutions in the two tracks, achieving impressive results on the V3Det dataset.  We look forward to seeing more innovative approaches in our future challenge tracks.
{
    \small
    \bibliographystyle{ieeenat_fullname}
    \bibliography{main}
}

\clearpage
\appendix

\section*{\centering Appendix}

\section{Teams and Affiliations}

\textbf{V3Det Challenge 2024}

\noindent \textbf{Organizer(s)}

\noindent Jiaqi Wang$^1$ (\texttt{wangjiaqi@pjlab.org.cn}), Yuhang Zang$^1$, Pan Zhang$^1$, Tao Chu$^1$, Yuhang Cao$^{1,2}$, Zeyi Sun$^{1,3}$, Ziyu Liu$^{1,4}$, Xiaoyi Dong$^{1,2}$, Tong Wu$^{2}$, Dahua Lin$^{1,2}$

\noindent \textbf{Affiliation(s)}

\noindent $^1$Shanghai AI Laboratory, $^2$Chinese University of Hong Kong, $^3$Shanghai Jiao Tong University, $^4$ Wuhan University

\hfill \break

\noindent \textbf{Team czm369}

\noindent \textbf{Title}

\noindent MixPL v2: Expanding Detection Categories via Semi-Supervised Learning

\noindent \textbf{Member(s)}

\noindent Zeming Chen$^{1}$ (\texttt{czm20@mails.tsinghua.edu.cn}), Zhi Wang$^{\dagger1}$

\noindent \textbf{Affiliation(s)}

\noindent $^1$Shenzhen International Graduate School, Tsinghua University

\hfill \break

\noindent \textbf{Team lcmeng}

\noindent \textbf{Title}

\noindent RichSem-DINO-FocalNet for V3Det Challenge 2024

\noindent \textbf{Member(s)}

\noindent Lingchen Meng$^{*1,2}$ (\texttt{lcmeng20@fudan.edu.cn}), Wenhao Yao$^{*1,2}$ (\texttt{whyao23@m.fudan.edu.cn}), Jianwei Yang$^{3}$, Sihong Wu$^{1,2}$, Zhineng Chen$^{1,2}$, Zuxuan Wu$^{1,2}$, Yu-Gang Jiang$^{1,2}$

\noindent \textbf{Affiliation(s)}

\noindent $^{1}$Shanghai Key Lab of Intell. Info. Processing, School of CS, Fudan University, $^{2}$Shanghai Collaborative Innovation Center of Intelligent Visual Computing, $^{3}$Microsoft Research, Redmond

\hfill \break

\noindent \textbf{Team TCSVT}

\noindent \textbf{Title}

\noindent Enhanced Object Detection: A Study on Vast Vocabulary Object Detection Track for V3Det Challenge 2024

\noindent \textbf{Member(s)}

\noindent Peixi Wu$^{*1}$ (\texttt{wupeixi@mail.ustc.edu.cn}), Bosong Chai$^{*\dagger2}$ (\texttt{chaibosong@mail.zju.edu.cn}), Xuan Nie$^{3}$, Longquan Yan$^{4}$, Zeyu Wang$^{2}$, Qifan Zhou$^{3}$, Boning Wang$^{2}$

\noindent \textbf{Affiliation(s)}

\noindent $^{1}$University of Science and Technology of China, $^{2}$Zhejiang University, $^{3}$Northwestern Polytechnical University, $^{4}$Northwest University

\hfill \break

\noindent \textbf{Team Innovision}

\noindent \textbf{Title}

\noindent 3rd Place Solution for OVD Track in CVPR 2024 VPLOW workshop: Visual Perception via Learning in an Open World

\noindent \textbf{Member(s)}

\noindent Jiaqi Huang$^{1}$ (\texttt{huangjq23@mails.tsinghua.edu.cn}), Zunnan Xu$^{1}$, Xiu Li$^{1}$, Kehong Yuan$^{\dagger1}$

\noindent \textbf{Affiliation(s)}

\noindent $^{1}$Tsinghua Shenzhen International Graduate School, Tsinghua University

\hfill \break

\noindent \textbf{Team JYYY}

\noindent \textbf{Title}

\noindent Team JYYY: V3Det Challenge 2024 - Vast Vocabulary Visual Detection

\noindent \textbf{Member(s)}

\noindent Yanyan Zu$^{1}$ (\texttt{23171214480@stu.xidian.edu.cn}), Jiayao Hao$^{1}$, Qiong Gao$^{1}$, Licheng Jiao$^{1}$

\noindent \textbf{Affiliation(s)}

\noindent $^{1}$Intelligent Perception and Image Understanding Lab, Xidian University

\end{document}